\documentclass[10pt,twocolumn,letterpaper]{article}

\usepackage{cvpr}              
\usepackage{amsmath}
\usepackage{amssymb}
\usepackage{mathtools}
\usepackage{amsthm}
\usepackage{bm}
\makeatletter
\renewcommand\paragraph{%
  \@startsection{paragraph}{4}{\z@}%
    {1.5ex \@plus0.5ex \@minus0.2ex}
    {-1em}
    {\normalfont\normalsize\bfseries}}
\makeatother
\usepackage{enumitem}
\usepackage{stfloats}
\usepackage{multirow}
\usepackage[table]{xcolor}
\definecolor{deepgreen}{HTML}{228B22} 

\definecolor{cvprblue}{rgb}{0.21,0.49,0.74}
\usepackage[pagebackref,breaklinks,colorlinks,allcolors=cvprblue]{hyperref}

\def\paperID{5147} 
\def\confName{CVPR}
\def\confYear{2026}

\title{BReD: BEV Residual Diffusion for Feed-Forward Road Gaussian Reconstruction}

\author{Baizhi Wang\\
Sun Yat-sen University\\
}

\begin{document}
\maketitle
\begin{abstract}
Accurate road surface reconstruction is essential for autonomous-driving mapping and simulation, but existing methods trade reconstruction quality for cross-scene inference. Optimization-based methods recover detailed surfaces through costly per-scene optimization, while feed-forward methods lack road-specific constraints and often produce large elevation errors. We propose \textbf{BReD}, a feed-forward framework for reconstructing road elevation, appearance, and semantics from calibrated six-camera sequences. BReD first estimates coarse elevation through LiDAR-supervised depth learning and BEV lifting, using only cameras at inference. A lightweight BEV residual diffusion model then conditions elevation refinement on a camera-derived depth--elevation discrepancy and elevation-aware image features. Diffusion-indexed perturbations expose the model to different discrepancy corruption levels during training, while inference performs one deterministic refinement from the clean discrepancy. The refined elevations are temporally aggregated to anchor road-aligned Gaussians, whose attributes are predicted by a geometry-prior-guided decoder without test-time optimization. On four held-out nuScenes scenes, BReD achieves 7.34 cm elevation RMSE and 90.92\% semantic mIoU, compared with 14.43 cm for RoGS without test-time LiDAR and 8.88 cm for its test-time-LiDAR-assisted variant; both RoGS variants require per-scene optimization.
\end{abstract}
    
\vspace{-5pt}
\section{Introduction}
\label{sec:intro}

Accurate road surface reconstruction is fundamental to autonomous driving, supporting localization, mapping, simulation, and closed-loop planning through a joint representation of metric geometry, appearance, and semantics. Road elevation is particularly important for distinguishing slopes, curbs, intersections, and subtle structures lost in planar bird's-eye-view (BEV) representations. Conventional high-definition (HD) map production typically relies on dedicated survey vehicles equipped with costly LiDAR systems, followed by substantial processing or annotation of road elements. A transferable camera-only inference pipeline could instead lower the sensing cost and enable scalable map collection from production vehicles or crowdsourced fleets, while jointly reconstructed semantics could serve as pre-annotations for downstream map production. Although surround-view cameras offer such a scalable and low-cost sensing setup, recovering renderable road surfaces from images remains difficult because perspective observations provide indirect geometry, distant regions contain limited visual evidence, and dynamic objects frequently occlude the road. Effective reconstruction must therefore infer fine-grained elevation while preserving appearance and semantics across the camera surround.
Existing methods mainly follow two paradigms. Optimization-based approaches represent roads with meshes or Gaussian primitives and optimize their geometry and attributes per sequence~\cite{mei2024rome,wu2024emie,deng2026rogs,peng2025magicroad}. They can produce detailed reconstructions, especially with dense LiDAR, but incur substantial test-time cost and cannot directly transfer learned road priors across scenes. Feed-forward geometry models instead support cross-scene inference~\cite{wang2025vggt,jia2025drivevggt,tian2025drivingforward}, but lack road-specific metric and topological constraints, often causing large vertical errors and rendering artifacts on non-planar roads. Recent road-specific feed-forward methods mainly target monocular or sparse-view inputs~\cite{wu2026bev,jiao2026roadvggt}. Joint feed-forward reconstruction of elevation, RGB, and semantics from six-camera observations remains insufficiently explored.
This setting presents three challenges. First, depth-based BEV lifting unifies surround-view features into a metric surface, but discretized frustum depths introduce quantization artifacts and lose centimeter-level height variations. Second, refining the coarse surface requires sampling image evidence near its estimated 3D location, where elevation errors can cause feature misalignment. Such refinement must exploit perspective cues and imperfect camera-derived geometry without ground-truth elevation at inference. Third, the BEV elevation must be converted into dense, renderable primitives with multi-view-consistent geometry, appearance, opacity, orientation, scale, and semantics, without resorting to per-scene optimization.

We address these challenges with \textbf{BReD}, a \textbf{B}EV \textbf{Re}sidual \textbf{D}iffusion framework for feed-forward road Gaussian reconstruction. BReD first predicts frame-wise coarse elevation through depth-supervised BEV lifting. It then introduces a conditional residual diffusion module that refines the coarse elevation rather than generating absolute elevation from pure noise. Specifically, the discrepancy between the coarse elevation and a camera-depth-derived elevation serves as an observable geometric condition, while elevation-aware image features provide local perspective evidence. Diffusion-indexed perturbations expose a timestep-conditioned ConvGRU to different discrepancy corruption levels during training; inference uses the clean discrepancy at $t=0$ for one deterministic refinement step.
The frame-wise refined elevations are transformed into a common reference frame and temporally aggregated to anchor dense road-aligned Gaussians. A geometry-prior-guided MLP predicts bounded residuals for elevation, color, opacity, scale, rotation, and semantics, producing a renderable sequence-level representation without test-time gradient optimization. The elevation generator and Gaussian decoder are trained in two stages.
We evaluate BReD on four held-out scenes from the official nuScenes validation split. Although trained with LiDAR-derived depth and elevation supervision, BReD uses only calibrated cameras and poses at inference and requires no scene-specific optimization. 
It achieves a mean elevation RMSE of $7.34\,\mathrm{cm}$, compared with $14.43\,\mathrm{cm}$ for RoGS without LiDAR and $8.88\,\mathrm{cm}$ for its LiDAR-assisted variant, together with $90.92\%$ semantic mIoU and $24.15\,\mathrm{dB}$ PSNR. These results indicate the potential of transferable camera-only inference for scalable geometric map reconstruction and semantic pre-annotation. On \texttt{scene-0523}, the elevation generation and Gaussian refinement stages progressively reduce RMSE from $19.60$ to $9.47$ and $5.07\,\mathrm{cm}$, respectively.
Our contributions are summarized as follows:
\begin{itemize}
    \item We introduce BReD, a feed-forward framework for scalable joint road elevation, appearance, and semantic reconstruction from calibrated six-camera sequences, requiring only cameras and no scene-specific optimization at inference.
    \item We formulate elevation refinement as conditional BEV residual diffusion, where a camera-derived depth--elevation discrepancy provides an observable residual condition, elevation-aware image features provide local perspective evidence, and diffusion-indexed perturbations expose a timestep-conditioned ConvGRU to different discrepancy corruption levels during training.
    \item We develop a geometry-prior-guided Gaussian decoder that converts temporally aggregated BEV surfaces into renderable road Gaussians; on four held-out nuScenes scenes, BReD achieves $7.34\,\mathrm{cm}$ elevation RMSE and $90.92\%$ mIoU.
\end{itemize}

\vspace{-3pt}
\section{Related Work}
\label{sec_related}

\vspace{-3pt}
\subsection{BEV Perception and Elevation Estimation}
\label{sec_related_bev}

\vspace{-3pt}
Camera-based BEV perception transforms perspective-view features into a metric top-down representation. Lift-Splat-Shoot~\cite{philion2020lift} predicts categorical depth distributions, lifts image features into camera frustums, and aggregates them on a BEV grid. BEVDepth~\cite{li2023bevdepth} introduces explicit LiDAR depth supervision and camera-aware depth prediction, while BEVStereo~\cite{li2023bevstereo} uses temporal stereo cues to improve depth estimation. Transformer-based methods such as BEVFormer~\cite{li2024bevformer} instead construct spatial and temporal BEV features through attention. These architectures have supported 3D detection, occupancy prediction, and semantic mapping~\cite{zhang2023occformer,zhou2023matrixvt}. 
RoadBEV~\cite{zhao2024roadbev} reconstructs road elevation and RGB appearance from monocular or stereo images, while FastRSR~\cite{zhao2025fastrsr} improves this setting through depth-aware multi-scale projection and elevation-distribution prediction. 
Their released methods and evaluation protocols are designed for monocular or stereo front-view inputs and cannot be directly applied to the more complex six-camera surround-view setting of autonomous-driving datasets such as nuScenes. In contrast, our method directly processes surround-view observations from autonomous-driving scenes and jointly reconstructs road elevation, RGB appearance, and semantics.

\vspace{-3pt}
\subsection{Optimization-Based Road Reconstruction}
\label{sec_related_optimization}
\vspace{-3pt}
Road surface reconstruction represents road geometry together with appearance or map-related attributes. RoMe~\cite{mei2024rome} models large road regions using an explicit triangular mesh, a learned elevation function, and per-vertex color and semantic attributes. 
EMIE-MAP~\cite{wu2024emie} combines an explicit mesh with implicit elevation and appearance encoding to model roads observed by surround-view cameras. RoGS~\cite{deng2026rogs} places 2D Gaussian surfels on a road-aligned meshgrid and optimizes their geometry, color, opacity, scale, rotation, and semantic attributes. Related driving-scene methods extend NeRF or Gaussian Splatting to street reconstruction, including static and dynamic scene components~\cite{kerbl20233d,zhou2024drivinggaussian,zhou2024hugs}. 
MagicRoad~\cite{peng2025magicroad} further incorporates semantic-guided obstacle inpainting and appearance enhancement into road reconstruction. These approaches obtain scene-specific road representations through optimization over the observations of each target sequence. They establish effective road representations and rendering objectives, while feed-forward approaches seek to reuse learned road priors across scenes.

\vspace{-3pt}
\subsection{Feed-Forward Geometry and Gaussian Reconstruction}
\label{sec_related_feedforward}
\vspace{-3pt}
Feed-forward reconstruction predicts scene geometry or renderable primitives directly from input images. PixelSplat~\cite{charatan2024pixelsplat} predicts pixel-aligned Gaussians from posed image pairs, while MVSplat~\cite{chen2024mvsplat} constructs Gaussian representations from sparse multi-view features and plane-sweep correlations. DrivingForward~\cite{tian2025drivingforward} extends this paradigm to flexible surround-view inputs for driving-scene novel-view synthesis. GaussianFormer~\cite{huang2024gaussianformer} uses Gaussian primitives for semantic occupancy prediction rather than surface rendering. Geometry foundation models provide a complementary representation: VGGT~\cite{wang2025vggt} jointly predicts camera parameters, depth, point maps, and tracks, and DriveVGGT~\cite{jia2025drivevggt} incorporates temporal and calibrated multi-camera constraints for driving sequences.
Road-specific feed-forward methods combine direct prediction with structured surface representations. BEV-GS~\cite{wu2026bev} predicts road elevation and texture from a single image and constructs grid Gaussians for real-time rendering in a compact front-road region. RoadVGGT~\cite{jiao2026roadvggt} uses a geometric foundation model to predict dense road Gaussians and consolidates them through confidence-weighted road-plane fusion, with category-aware processing around road structures. Its semantic maps are provided as inputs and associated with the reconstructed Gaussians, whereas its learned Gaussian attributes cover geometry, appearance, opacity, scale, and rotation. BReD follows the feed-forward road Gaussian setting but centers its design on surround-view BEV elevation generation: a residual diffusion module refines the coarse elevation, and a geometry-prior-guided residual head jointly predicts Gaussian geometry, appearance, opacity, and semantic logits. This design connects depth-supervised BEV lifting, fine-grained elevation refinement, and renderable road reconstruction within one cross-scene inference pipeline.
\begin{figure*}[!t]
\centering
\includegraphics[width=0.86\textwidth]{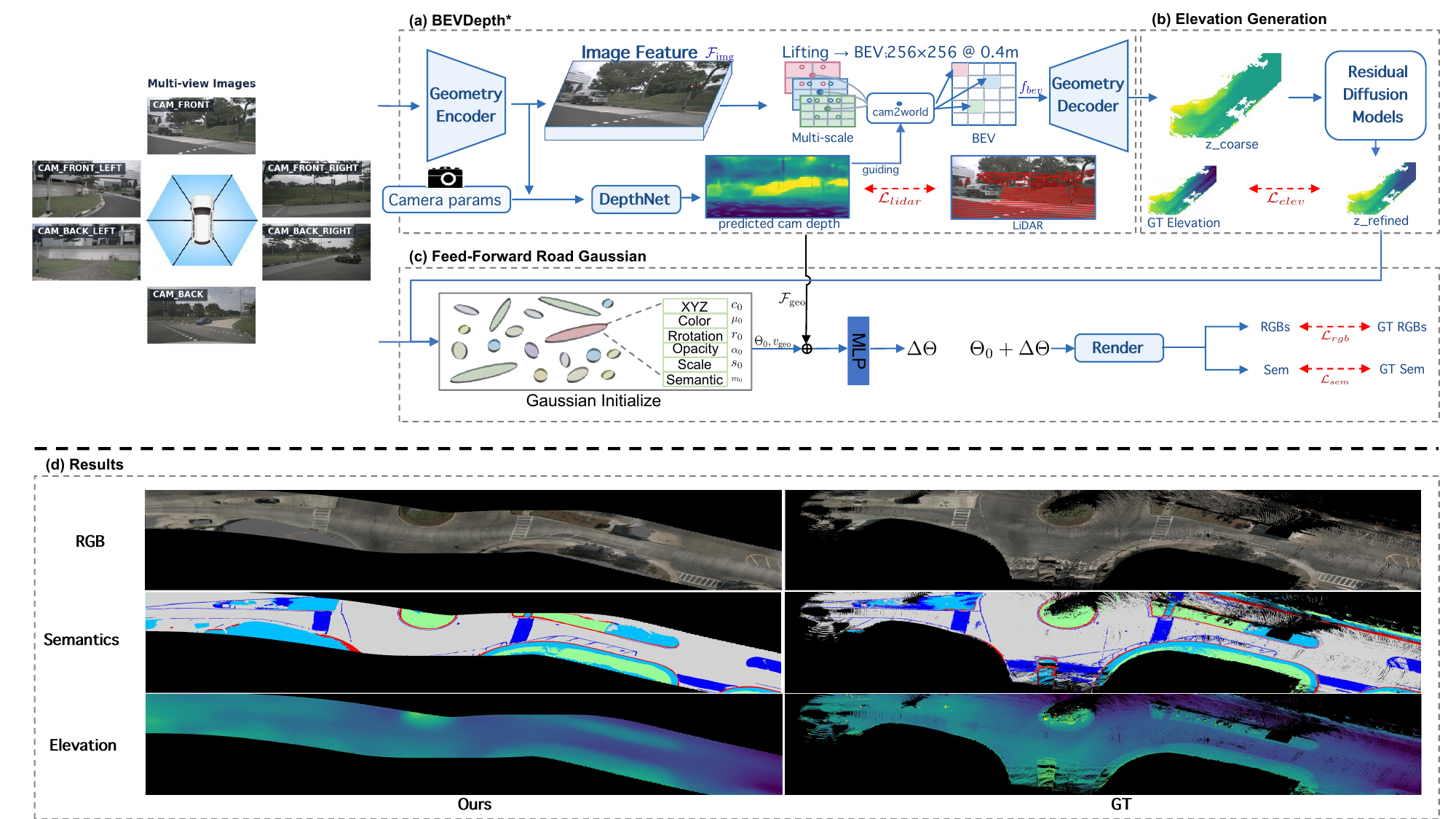}
\caption{\textbf{BReD overview.}
\textbf{(a)} \texttt{BEVDepth*} predicts frame-wise coarse elevation using LiDAR depth supervision during training and camera-predicted depth at inference.
\textbf{(b)} \texttt{ReD*} combines depth--elevation discrepancy, BEV context, and elevation-aware image features for single-step residual denoising.
\textbf{(c)} Refined elevations are temporally aggregated, and \texttt{FFGS} predicts road-Gaussian attributes without test-time optimization.
\textbf{(d)} Reconstructed RGB, semantics, and elevation.}
\label{fig:overview}
\vspace{-5pt}
\end{figure*}

\vspace{-5pt}
\section{Methodology}
\label{sec:method}
\vspace{-3pt}
BReD contains three stages (Fig.~\ref{fig:overview}):
1) \textbf{\texttt{BEVDepth*}} predicts frame-wise coarse elevation $z_{\text{coarse}}$ through depth-guided BEV lifting;
2) \textbf{\texttt{ReD*}} performs single-step residual denoising conditioned on depth--elevation discrepancy, BEV context, and elevation-aware features; and
3) \textbf{\texttt{FFGS}} temporally aggregates refined surfaces and predicts bounded Gaussian attribute residuals without test-time gradient optimization.

\vspace{-3pt}
\subsection{BEVDepth: Coarse Elevation Initialization}
\label{subsec:bevdepth}
\vspace{-3pt}
Given $N=6$ surround-view images $\mathcal{I} = \{I_i\}_{i=1}^6$ and their camera parameters $\mathcal{P} = \{K_i, T_i\}_{i=1}^6$, we adapt the BEVDepth~\cite{li2023bevdepth} lifting architecture as our foundational module (Fig.~\ref{fig:overview}(a)). A Geometry Encoder (ResNet-50) extracts hierarchical perspective features, where intermediate multi-scale features $\{P_2, P_3, P_4\}$ are retained to construct the multi-scale elevation embedding $f_{ms}$ in the diffusion condition features (Fig.~\ref{fig:ms_embed}), and the top-level $P_5$ feature serves as the primary representation $\mathcal{F}_{\text{img}}$. Concurrently, $\mathcal{F}_{\text{img}}$ and $\mathcal{P}$ are fed into a camera-aware \textbf{DepthNet}~\cite{kumar2018depthnet} to predict pixel-wise frustum depth distributions $D_{\text{pred}}$, directly supervised by projected LiDAR points via $\mathcal{L}_{lidar}$.

Guided by $D_{\text{pred}}$ and camera-to-world (\texttt{cam2world}) transformations $T_i$, $\mathcal{F}_{\text{img}}$ is lifted into 3D frustum points and splatted onto a $256 \times 256$ BEV grid ($0.4\text{m} \times 0.4\text{m}$ cell size), producing the aggregated BEV feature map $f_{bev} \in \mathbb{R}^{C \times 256 \times 256}$. To adapt the original BEVDepth from 3D object detection to road surface reconstruction, we replace its detection heads with a lightweight \textbf{Geometry Decoder} $\Phi_{\text{dec}}$ and modify the final output layer to regress a 1D continuous BEV elevation map $z_{\text{coarse}} \in \mathbb{R}^{1 \times 256 \times 256}$ (denoted as \texttt{BEVDepth*}):
\begin{equation}
    f_{bev} = \text{Lift}\left(\mathcal{F}_{\text{img}}, D_{\text{pred}}, \mathcal{P}\right), \quad z_{\text{coarse}} = \Phi_{\text{dec}}\left(f_{bev}\right).
\end{equation}

\begin{figure}[t]
    \centering
    \includegraphics[width=\linewidth]{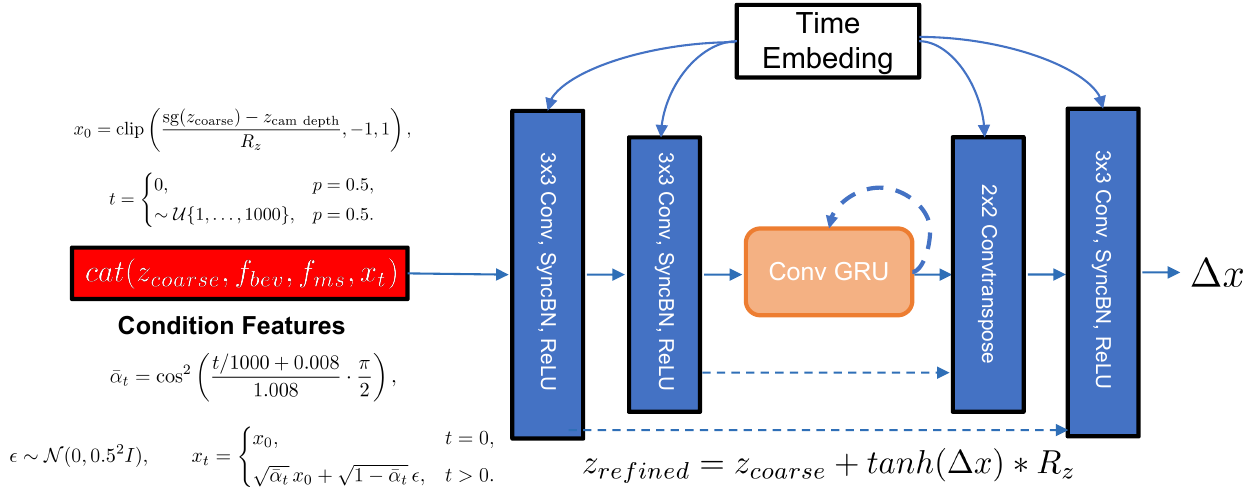}
    \caption{\textbf{BEV Residual Diffusion.}
During training, the camera-derived discrepancy $x_0$ is retained at $t=0$ or perturbed at $t>0$. A timestep-conditioned \texttt{ConvGRU} predicts a bounded elevation update. Inference uses the clean $t=0$ state in one denoising step.}
    \label{fig:red_arch}
\end{figure}
\vspace{-3pt}
\subsection{Elevation Generation via Residual Diffusion}
\label{subsec:red}
\begin{figure}[t]
    \centering
    \includegraphics[width=0.8\linewidth]{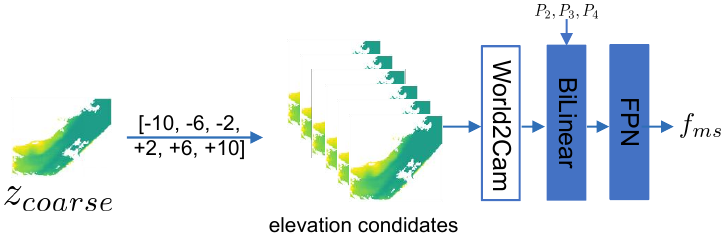}
    \caption{\textbf{Extraction of the Multi-Scale Elevation Embedding ($f_{ms}$).} 
    Centered around $z_{\text{coarse}}$, vertical elevation candidates with offsets $[-10, -6, -2, +2, +6, +10]\,\text{cm}$ are projected onto multi-view image planes via \texttt{World2Cam}. Multi-scale features $\{P_2, P_3, P_4\}$ are bilinearly sampled (\texttt{BiLinear}) and aggregated through an \texttt{FPN} block to produce $f_{ms}$.}
    \label{fig:ms_embed}
\end{figure}

\vspace{-3pt}
While $z_{\text{coarse}}$ captures the global road slope, direct regression from discretized frustum depths inevitably suffers from quantization artifacts. To recover fine-grained road topography, we introduce the \textbf{Elevation Generation} module driven by a lightweight \textbf{BEV Residual Diffusion Model} (Fig.~\ref{fig:overview}(b)), whose internal workflow and multi-scale feature extraction are illustrated in Fig.~\ref{fig:red_arch} and Fig.~\ref{fig:ms_embed}.

\noindent\textbf{Multi-Scale Elevation Embedding ($f_{ms}$).}
To provide the diffusion denoiser with fine-grained perspective cues around the initial surface estimate, we construct an elevation-aware multi-scale feature embedding $f_{ms}$ (Fig.~\ref{fig:ms_embed}). Using $z_{\text{coarse}}$ as a spatial anchor, we generate $L=6$ discrete elevation candidates by sweeping centimeter-level vertical offsets $\Delta h \in \{-10, -6, -2, +2, +6, +10\}\,\text{cm}$ around $z_{\text{coarse}}$. These 3D candidate volumes are projected onto the surround-view image planes via world-to-camera (\texttt{World2Cam}) transformations for differentiable bilinear sampling (\texttt{BiLinear}) across multi-scale feature levels $\{P_2, P_3, P_4\}$, which are subsequently fused by an \texttt{FPN} module to yield $f_{ms}$.

\noindent\textbf{Conditional Residual Formulation.}
We refine an initial geometry through a diffused residual condition rather than generating the complete elevation map from pure noise. Given the image-derived features and two camera-derived elevation estimates, the desired road surface can be viewed as the conditional MAP estimate
\begin{equation}
    z^{*}=\arg\max_{z}\,
    p\!\left(z\mid f_{bev},f_{ms},z_{\text{coarse}},
    z_{\text{cam\_depth}}\right).
\end{equation}
This posterior combines image evidence with a road-geometry prior centered on $z_{\text{coarse}}$. We encode the disagreement between the two camera-derived estimates as an observable residual condition and train the refiner across multiple corruption levels, enabling it to recover a deterministic elevation correction from imperfect camera geometry.

\noindent\textbf{Normalized Discrepancy Prior and Hybrid Forward Diffusion.}
As depicted in Fig.~\ref{fig:red_arch}, we project $D_{\text{pred}}$ onto the $256 \times 256 \text{ @ } 0.4\text{m}$ BEV grid to obtain a depth-derived elevation reference $z_{\text{cam\_depth}}$, and formulate the normalized initial elevation discrepancy $x_0 \in [-1,1]^{1\times256\times256}$ as
\begin{equation}
    x_0 = \text{clip}\left(
    \frac{\text{sg}(z_{\text{coarse}})-z_{\text{cam\_depth}}}{R_z},
    -1,1\right),
\end{equation}
where $\text{sg}(\cdot)$ denotes stop-gradient and $R_z=0.3\text{m}$. Unlike a ground-truth residual, $x_0$ is an observable geometric condition formed entirely from camera-derived quantities and is therefore available during both training and inference.
We construct a diffusion-indexed family of discrepancy states by training on both clean and perturbed inputs. With probability $0.5$, we set $t=0$ and $x_t=x_0$; otherwise, we sample $t\sim\mathcal{U}\{1,\ldots,1000\}$ and perturb $x_0$ using a cosine diffusion schedule:
\begin{equation}
    x_t = \begin{cases}
    x_0, & t = 0, \\
    \sqrt{\bar{\alpha}_t}\,x_0
    + \sqrt{1-\bar{\alpha}_t}\,\epsilon, & t > 0,
    \end{cases}
    \qquad \epsilon\sim\mathcal{N}(0,I).
\end{equation}
Rather than modeling an unconditional elevation distribution, this forward process exposes a shared conditional refiner to different corruption levels of the camera-derived discrepancy.

\noindent\textbf{ConvGRU-Based Conditional Denoising.}
We construct the condition features by concatenating the coarse elevation $z_{\text{coarse}}$, BEV features $f_{bev}$, multi-scale elevation embeddings $f_{ms}$, and discrepancy state $x_t$ along the channel dimension (denoted as \texttt{ReD*} when incorporating $f_{ms}$, and \texttt{ReD} without $f_{ms}$):
\begin{equation}
    F_{\text{cond}} = \text{cat}\left(
    z_{\text{coarse}},f_{bev},f_{ms},x_t\right).
\end{equation}
Conditioned on $F_{\text{cond}}$ and the sinusoidal time embedding of $t$, a lightweight recurrent encoder--decoder equipped with Convolutional Gated Recurrent Units (\textbf{ConvGRU}) predicts an unbounded residual update $\Delta x$. The refined BEV elevation is recovered as
\begin{equation}
    z_{\text{refined}}
    =z_{\text{coarse}}+\tanh(\Delta x)\cdot R_z.
\end{equation}
For every sampled timestep, $z_{\text{refined}}$ is directly supervised by $z_{\text{gt}}$. This direct elevation objective trains the network to infer the desired correction from discrepancy observations at different corruption levels; it does not require an explicit noise-prediction objective. Because $z_{\text{coarse}}$ and $x_0$ already provide an informative initialization, inference uses the clean discrepancy at $t=0$ and performs one deterministic refinement without noise injection, iterative reverse sampling, or test-time optimization.

\subsection{Feed-Forward Road Gaussian}
\label{subsec:road_gaussian}
In \texttt{FFGS}, frame-wise refined elevations are transformed into a common reference frame and temporally aggregated to initialize road Gaussians, whose attribute residuals are predicted by a geometry-prior-guided MLP without test-time gradient optimization.

\noindent\textbf{Gaussian and Geometry Prior Initialization.} 
We discretize the road corridor (within $8\text{m}$ of the vehicle trajectory) into a $0.05\text{m}$ BEV grid of $M$ vertices with fixed horizontal coordinates $\mu_{xy} = (x_0, y_0)$. Frame-wise $0.4\text{m}$ refined elevations $z_{\text{refined}}$ are transformed into the reference frame, queried within a $0.4/\sqrt{2}\text{m}$ radius ($i.e.$, the center-to-corner half-diagonal of a $0.4\text{m}$ square BEV cell for seamless coverage), and aggregated via the temporal median to initialize vertex elevations $z_0 \in \mathbb{R}^{M \times 1}$ (uncovered vertices fallback to the nearest trajectory height). Projecting the 3D anchors $\mu_0 = [x_0, y_0, z_0]^T$ (\texttt{XYZ}) onto observed sequence views $i$ with valid depth $d_i \in (1\text{m}, 20\text{m})$ and road mask ($\ge 0.5$), we compute the multi-view median RGB probabilities $p_{\text{rgb}}$ and map them into logit space via $c_0 = \log(p_{\text{rgb}} / (1 - p_{\text{rgb}}))$ (\texttt{Color}). Vertical scales are fixed to zero ($s_z = 0$), while horizontal scale logits $s_0 \in \mathbb{R}^{M \times 2}$ (\texttt{Scale}), trajectory-aligned quaternion rotations $r_0 \in \mathbb{R}^{M \times 4}$ (\texttt{Rotation}), logit opacities $\alpha_0 \in \mathbb{R}^{M \times 1}$ (\texttt{Opacity}), and semantic logits $m_0 \in \mathbb{R}^{M \times C_{\text{sem}}}$ (\texttt{Semantic}) are initialized to canonical values, forming the initial parameter set $\Theta_0 = \{z_0, s_0, r_0, \alpha_0, c_0, m_0\}$. Concurrently, we reuse the frozen multi-scale elevation embeddings $f_{ms}$ ($32 \times 3 = 96$-D across $\{P_2, P_3, P_4\}$, Fig.~\ref{fig:ms_embed}), query them onto the Gaussian vertices within a $0.8/\sqrt{2}\text{m}$ radius ($2\times$ the cell half-diagonal) via multi-frame mean pooling, and concatenate them with $z_0$ to form the $97$-D \textbf{Geometry Prior} $\mathcal{F}_{\text{geo}} \in \mathbb{R}^{M \times 97}$. A binary coverage mask $v_{\text{geo}} \in \{0, 1\}^{M \times 1}$ is simultaneously recorded ($v_{\text{geo}} = 1$ if at least one valid BEV cell falls within the query radius, and $0$ otherwise) to indicate prior availability.

\noindent\textbf{MLP-Based Conditional Residual Refinement.} 
To prevent unaligned feature noise at uncovered vertices ($v_{\text{geo}} = 0$), we gate the geometry prior via $v_{\text{geo}} \odot \mathcal{F}_{\text{geo}}$ to strictly zero out invalid prior entries. Conditioned on the initial Gaussian attributes $\Theta_0$ and the gated geometry prior, a lightweight multi-layer perceptron (MLP) extracts a latent representation $\mathbf{h}$ and employs three task-specific linear heads ($\Phi_{\text{geo}}, \Phi_{\text{app}}, \Phi_{\text{sem}}$) to predict bounded attribute residuals in a single forward pass:
\begin{equation}
\begin{aligned}
    \mathbf{h} &= \text{MLP}\bigl(\text{cat}(\Theta_0, \, v_{\text{geo}} \odot \mathcal{F}_{\text{geo}})\bigr), \\
    \Delta \Theta &= \gamma \odot \tanh\bigl( [\Phi_{\text{geo}}(\mathbf{h}), \, \Phi_{\text{app}}(\mathbf{h}), \, \Phi_{\text{sem}}(\mathbf{h})] \bigr), \\
    \Theta_{\text{final}} &= \Theta_0 + \Delta \Theta,
\end{aligned}
\end{equation}
where $\Theta = \{z, s, r, \alpha, c, m\}$ and the residual bounds are set to $\gamma_z = 0.05\text{m}$, $\gamma_s = \gamma_r = 0.1$, and $\gamma_{\alpha} = \gamma_c = \gamma_m = 1.0$. Standard activations ($\text{Sigmoid}$ for $c, \alpha$; $\exp$ for $s$; $L_2$ normalization for $r$; $\text{Softmax}$ for $m$) are subsequently applied to $\Theta_{\text{final}}$ before differentiable rasterization (\texttt{Render}).

\vspace{-3pt}
\subsection{Training Objectives}
\label{subsec:training_objectives}
\vspace{-3pt}
We adopt a two-stage training paradigm to sequentially optimize the BEV elevation generation modules and the feed-forward road Gaussian refiner.

\noindent\textbf{Stage 1: BEV Elevation Generation.} 
In Stage 1, we jointly optimize \texttt{BEVDepth*} and \texttt{ReD*} with unit loss weights. \textbf{DepthNet} discretizes the camera frustum into $B$ depth bins and is supervised by one-hot LiDAR depth projections $D_{\text{gt}}$ over valid LiDAR pixels $\Omega_{\text{lidar}}$ via Binary Cross-Entropy (BCE):
\begin{equation}
    \mathcal{L}_{lidar} = \frac{1}{|\Omega_{\text{lidar}}|} \sum_{p \in \Omega_{\text{lidar}}} \text{BCE}\bigl(D_{\text{pred}}(p), \, D_{\text{gt}}(p)\bigr).
\end{equation}
Meanwhile, both $z_{\text{coarse}}$ and $z_{\text{refined}}$ are supervised against the ground-truth road elevation $z_{\text{gt}}$ over valid BEV road cells $\Omega_{\text{bev}}$ via an $\mathcal{L}_1$ regression loss:
\begin{equation}
\begin{aligned}
    \mathcal{L}_{elev} = \frac{1}{|\Omega_{\text{bev}}|} \sum_{u \in \Omega_{\text{bev}}} \Bigl( 
    &\bigl| z_{\text{coarse}}(u) - z_{\text{gt}}(u) \bigr| \\
    &+ \bigl| z_{\text{refined}}(u) - z_{\text{gt}}(u) \bigr| \Bigr).
\end{aligned}
\end{equation}
The overall Stage 1 objective is $\mathcal{L}_{\text{stage1}} = \mathcal{L}_{lidar} + \mathcal{L}_{elev}$.

\begin{table*}[!t]
  \caption{Results on four held-out nuScenes scenes under the same valid road support and Mask2Former-derived semantic reference. BReD performs camera-only feed-forward inference; RoMe and RoGS use per-scene optimization, with target-sequence LiDAR used only by RoGS w/ LiDAR.}
  \label{tab:metric}
  \centering
  \small
  \setlength{\tabcolsep}{6.5pt}
  \renewcommand{\arraystretch}{1.08}
  \newcommand{\muted}[1]{\textcolor{black!62}{#1}}

  \resizebox{0.76\textwidth}{!}{%
  \begin{tabular}{@{}lllccccc@{}}
    \toprule
    \multirow{2}{*}{Method}
      & \multirow{2}{*}{Inference setting}
      & \multirow{2}{*}{Metric}
      & \multicolumn{5}{c}{Scenes} \\
    \cmidrule(lr){4-8}
      & & & 0035 & 0519 & 0523 & 0905 & Mean \\
    \midrule

    \multirow{3}{*}{RoMe~\cite{mei2024rome}}
      & \multirow{3}{*}{Camera, optimization}
      & PSNR$\uparrow$
      & 25.25 & 25.99 & 27.89 & 28.61 & 26.94 \\
      & & mIoU$\uparrow$
      & 79.91 & 85.33 & 85.90 & 86.26 & 84.35 \\
      & & Elev.$\downarrow$
      & 19.15 & 11.16 & 10.90 & 13.19 & 13.60 \\\midrule
    \addlinespace[3pt]

    \multirow{3}{*}{\muted{RoGS w/ LiDAR~\cite{deng2026rogs}}}
      & \multirow{3}{*}{\muted{Camera + LiDAR, optimization}}
      & \muted{PSNR$\uparrow$}
      & \muted{24.09} & \muted{22.71} & \muted{25.55}
      & \muted{22.75} & \muted{23.77} \\
      & & \muted{mIoU$\uparrow$}
      & \muted{82.67} & \muted{93.76} & \muted{91.10}
      & \muted{91.91} & \muted{89.86} \\
      & & \muted{Elev.$\downarrow$}
      & \muted{11.10} & \muted{7.10} & \muted{6.70}
      & \muted{10.60} & \muted{8.88} \\\midrule
    \addlinespace[3pt]

    \multirow{3}{*}{RoGS w/o LiDAR~\cite{deng2026rogs}}
      & \multirow{3}{*}{Camera, optimization}
      & PSNR$\uparrow$
      & 23.22 & 21.77 & 24.66 & 21.48 & 22.78 \\
      & & mIoU$\uparrow$
      & 75.29 & 90.91 & 87.08 & 87.21 & 85.12 \\
      & & Elev.$\downarrow$
      & 21.89 & 10.99 & 12.84 & 13.99 & 14.43 \\\midrule
    \addlinespace[3pt]

    \multirow{3}{*}{VGGT-Long~\cite{deng2026vggt,wang2025vggt}}
      & \multirow{3}{*}{Camera, feed-forward}
      & PSNR$\uparrow$
      & 14.27 & 14.74 & 16.74 & 14.57 & 15.08 \\
      & & mIoU$\uparrow$
      & 70.04 & 81.76 & 82.06 & 78.30 & 78.04 \\
      & & Elev.$\downarrow$
      & 28.20 & 48.80 & 35.50 & 39.30 & 37.95 \\\midrule
    \addlinespace[3pt]

    \multirow{3}{*}{DriveVGGT~\cite{jia2025drivevggt}}
      & \multirow{3}{*}{Camera, feed-forward}
      & PSNR$\uparrow$
      & 14.67 & 13.04 & 16.22 & 15.80 & 14.93 \\
      & & mIoU$\uparrow$
      & 74.65 & 84.35 & 81.84 & 83.92 & 81.19 \\
      & & Elev.$\downarrow$
      & 40.10 & 32.10 & 31.70 & 29.80 & 33.43 \\

    \midrule
    \multirow{3}{*}{\textbf{BReD (Ours)}}
      & \multirow{3}{*}{Camera, feed-forward}
      & PSNR$\uparrow$
      & 25.08 & 22.75 & 25.11 & 23.67 & 24.15 \\
      & & mIoU$\uparrow$
      & \textbf{86.99} & \textbf{94.13} & \textbf{91.16}
      & \textbf{91.39} & \textbf{90.92} \\
      & & Elev.$\downarrow$
      & \textbf{6.74} & \textbf{6.98} & \textbf{5.07}
      & \textbf{10.55} & \textbf{7.34} \\
    \bottomrule
  \end{tabular}}
\end{table*}
\noindent\textbf{Stage 2: Feed-Forward Road Gaussian Refinement.}
We freeze Stage 1 and train a shared Gaussian residual head across training sequences using aggregated elevation and geometry priors. At inference, the learned head is directly applied after deterministic multi-frame aggregation; no scene-specific parameters are optimized. To suppress non-road outliers, we construct a binary valid mask $\mathcal{M}(p) \in \{0, 1\}$ over the image domain $\Omega$ by intersecting the valid rendered depth range $d(p) \in (1\text{m}, 20\text{m})$ with the road mask, followed by $5 \times 5$ morphological erosion and dilation.

The exposure-calibrated rendered image $\hat{I}$ and semantic probabilities $\hat{S}$ are supervised against the observed image $I_{\mathrm{gt}}$ and one-hot Mask2Former semantic pseudo-labels $S_{\mathrm{gt}}$ across $C_{\mathrm{sem}}$ classes via masked photometric $L_1$ and cross-entropy losses:
\begin{equation}
    \mathcal{L}_{rgb} = \frac{1}{3|\Omega|} \sum_{p \in \Omega} \mathcal{M}(p) \, \bigl\| \hat{I}(p) - I_{\text{gt}}(p) \bigr\|_1,
\end{equation}
\begin{equation}
    \mathcal{L}_{sem} = -\frac{1}{|\Omega|} \sum_{p \in \Omega} \mathcal{M}(p) \sum_{c=1}^{C_{\text{sem}}} S_{\text{gt}}^{(c)}(p) \log \hat{S}^{(c)}(p).
\end{equation}
Furthermore, to enforce local topological continuity on the final refined elevation $z_{\text{final}}$, we impose a 4-connected grid neighborhood ($\mathcal{N}_4$) smoothness regularization across visible Gaussian primitives $\mathcal{V}_{\text{vis}}$:
\begin{equation}
    \mathcal{L}_{\text{smooth}} = \sum_{j \in \mathcal{V}_{\text{vis}}} \frac{1}{4} \sum_{n \in \mathcal{N}_4(j)} \bigl( z_{\text{final}, j} - z_{\text{final}, n} \bigr)^2.
\end{equation}
The overall Stage 2 objective is formulated as:
\begin{equation}
    \mathcal{L}_{\text{stage2}} = \mathcal{L}_{rgb} + \lambda_{\text{sem}} \mathcal{L}_{sem} + \lambda_{\text{smooth}} \mathcal{L}_{\text{smooth}},
\end{equation}
where the balancing weights are set to $\lambda_{\text{sem}} = 0.06$ and $\lambda_{\text{smooth}} = 0.003$.

\vspace{-3pt}
\section{Experiments}
\label{sec:experiments}
\vspace{-3pt}
\subsection{Experimental Setup}
\vspace{-3pt}
\noindent\textbf{Datasets.}
We train on the official $700$-scene nuScenes training split ($28{,}130$ keyframes) and evaluate on four held-out validation scenes. These include \texttt{scene-0523}, used by RoMe and RoGS, two urban intersection scenes (\texttt{scene-0035}, \texttt{scene-0519}), and one rainy sharp-turn scene (\texttt{scene-0905}). 
Following RoMe~\cite{mei2024rome} and RoGS~\cite{deng2026rogs}, all methods use the same evaluation scenes and seven-class road masks and semantic pseudo-labels generated by an off-the-shelf Mask2Former~\cite{cheng2021maskformer}.

\noindent\textbf{Baselines.}
We compare with RoMe~\cite{mei2024rome}, RoGS~\cite{deng2026rogs}, VGGT-Long~\cite{deng2026vggt,wang2025vggt}, and DriveVGGT~\cite{jia2025drivevggt}. RoMe and RoGS are run using their official implementations and optimized independently on each target sequence for two epochs, using the same camera frames, poses, and Mask2Former road masks as BReD. RoGS is evaluated both with and without target-sequence LiDAR; only RoGS w/ LiDAR uses LiDAR at inference. VGGT-Long uses the pretrained VGGT model with chunk acceleration, and DriveVGGT uses its official nuScenes-trained checkpoint; both receive the same calibrated camera sequences and perform feed-forward inference without per-scene optimization. BReD is trained with LiDAR supervision but uses only calibrated cameras and poses at inference, without target-sequence optimization.

\noindent\textbf{Implementation Details.}
We train BReD for $100$ epochs using six surround-view cameras. In Stage 1, the BEVDepth backbone and elevation head are optimized with AdamW at learning rates of $1\times10^{-5}$ and $2\times10^{-4}$, respectively, with a weight decay of $10^{-2}$. In Stage 2, we freeze the elevation network and geometry priors, and optimize the exposure model and residual head with Adam at $1\times10^{-3}$. Training uses eight NVIDIA RTX 3090 GPUs, while reconstruction and evaluation use one.

\noindent\textbf{Metrics.}
Following RoGS~\cite{deng2026rogs}, we construct the BEV reference by aggregating multi-frame LiDAR road scans, projecting them onto the temporally nearest camera views, averaging overlapping RGB observations, taking the mode of the Mask2Former pseudo-labels, and orthographically rasterizing the results. All methods are evaluated on the same valid road support, defined by the intersection of the rasterized LiDAR-covered region and the common road mask. We report: 1) PSNR ($\uparrow$), computed between rendered and reference RGB over this support; 2) mIoU ($\uparrow$), computed against the Mask2Former-derived semantic reference over seven categories defined by RoMe and RoGS~\cite{deng2026rogs,mei2024rome}: background, road, curb, lane marking, crosswalk, bike lane, and manhole; and 3) Elev. ($\downarrow$), the elevation RMSE in cm after matching each Gaussian center to its nearest reference LiDAR point within a 0.1\,m horizontal radius over the common valid support.

\begin{figure*}[htp]
    \centering
    \includegraphics[width=0.8\textwidth]{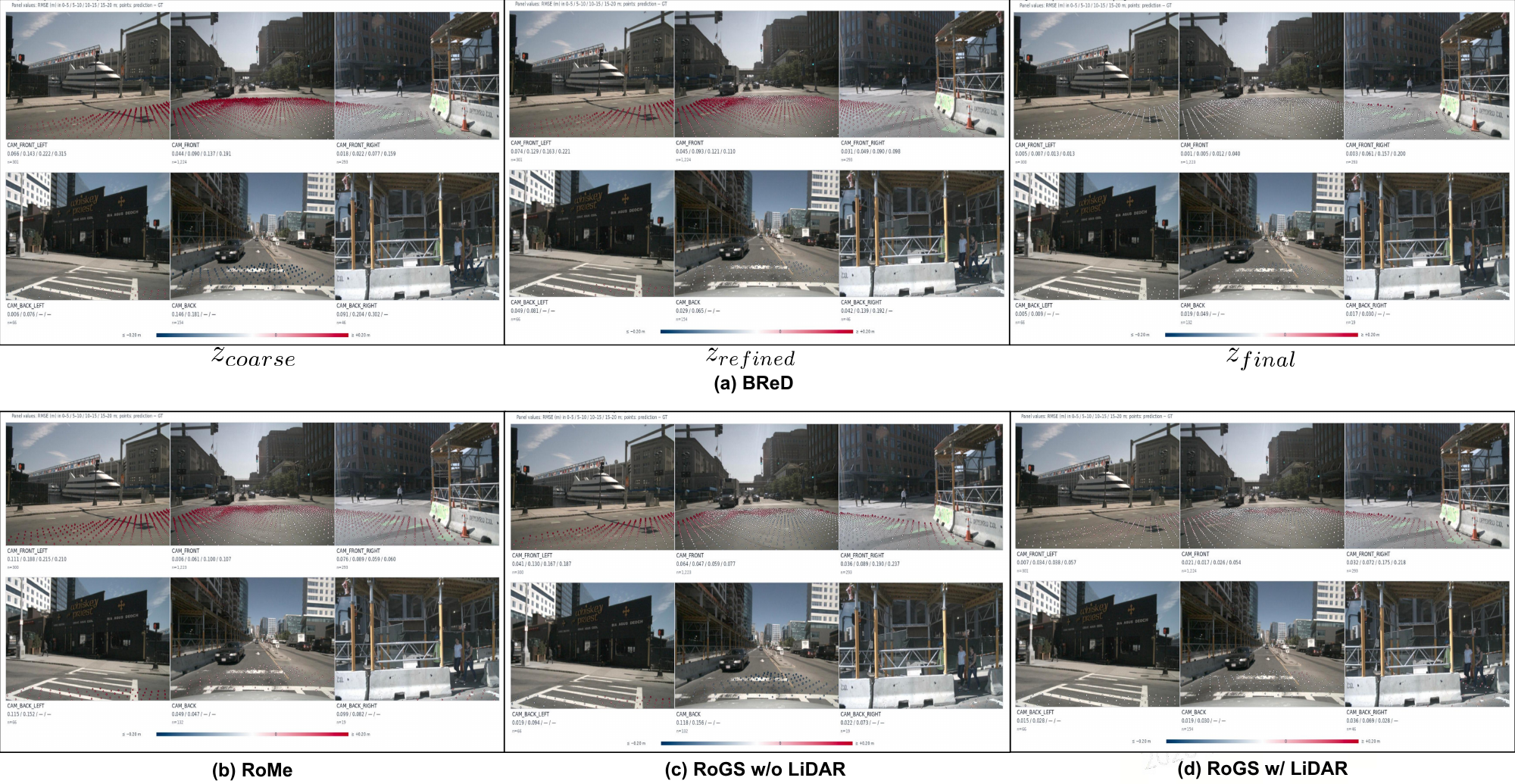}
    \caption{\textbf{Surround-view elevation errors.}
Results for (a) BReD, (b) RoMe, (c) RoGS w/o LiDAR, and (d) RoGS w/ LiDAR. Colors show signed vertical errors; statistics report RMSE and valid counts in $5\,\mathrm{m}$ distance bins.}
    \label{fig:elevation_projection}
\end{figure*}
\begin{table}[htp]
  \caption{Component-wise ablation study of BReD on road elevation accuracy. ``$\Delta$'' indicates the relative error reduction compared to the \texttt{BEVDepth*} baseline. \texttt{ReD} and \texttt{ReD*} denote the residual diffusion module without and with the multi-scale elevation embedding $f_{ms}$ in the condition features, respectively.}
  \label{tab:ablation}
  \centering
  \small
  \setlength{\tabcolsep}{2.5pt}
  \resizebox{0.8\columnwidth}{!}{
  \begin{tabular}{cccc|c|cc}
    \toprule
    \texttt{BEVDepth*} & \texttt{ReD} & \texttt{ReD*} & \texttt{FFGS} & Output & Elev.$\downarrow$ & $\Delta$ ($\%$) \\
    \midrule
    \checkmark & & & & $z_{\text{coarse}}$ & 19.60 & -- \\
    \checkmark & \checkmark & & & $z_{\text{refined}}$ & 16.49 & $-15.9\%$ \\
    \checkmark & & \checkmark & & $z_{\text{refined}}$ & 9.47 & $-51.7\%$ \\
    \checkmark & & \checkmark & \checkmark & $z_{\text{final}}$ & \textbf{5.07} & $\mathbf{-74.1\%}$ \\
    \bottomrule
  \end{tabular}
  }
\end{table}
\vspace{-5pt}
\subsection{Results on nuScenes}
\label{subsec:results_nuscenes}
\vspace{-3pt}
Table~\ref{tab:metric} compares methods on four representative nuScenes validation sequences. BReD, VGGT-Long, and DriveVGGT perform camera-only feed-forward inference, whereas RoMe and RoGS use per-scene optimization; RoGS w/ LiDAR additionally uses target-sequence LiDAR. RoGS (w/ LiDAR), highlighted in gray, uses dense LiDAR point clouds for per-scene optimization as a geometric reference.

\noindent\textbf{Visual and Semantic Quality.}
BReD achieves the highest mean mIoU ($90.92\%$) and a competitive PSNR ($24.15\,\mathrm{dB}$). RoMe obtains the highest PSNR through per-sequence optimization, while BReD better preserves semantics and metric geometry without test-time optimization.

\noindent\textbf{Elevation Accuracy.}
BReD achieves $7.34\,\mathrm{cm}$ mean RMSE, outperforming RoMe ($13.60\,\mathrm{cm}$), RoGS without LiDAR ($14.43\,\mathrm{cm}$), and the evaluated general feed-forward methods. On the same evaluation scenes, valid support, and metrics, it also obtains lower error than the per-scene-optimized RoGS variant using target-sequence LiDAR (8.88 cm).

\vspace{-4pt}
\subsection{Ablation Studies}
\vspace{-3pt}
\label{subsec:ablation}
Table~\ref{tab:ablation} evaluates the three stages on \texttt{scene-0523}. \texttt{BEVDepth*} obtains $19.60\,\mathrm{cm}$ RMSE; residual denoising without $f_{ms}$ reduces it to $16.49\,\mathrm{cm}$, while elevation-aware image features in \texttt{ReD*} further reduce it to $9.47\,\mathrm{cm}$. Temporal aggregation and Gaussian refinement in FFGS further reduce the RMSE to 5.07 cm, confirming the complementary effects of metric lifting, elevation-aware refinement, and Gaussian decoding.

\vspace{-3pt}
\subsection{Discussion}
\label{subsec:discussion}
\begin{figure}[htp]
  \centering
  \includegraphics[width=0.8\columnwidth]{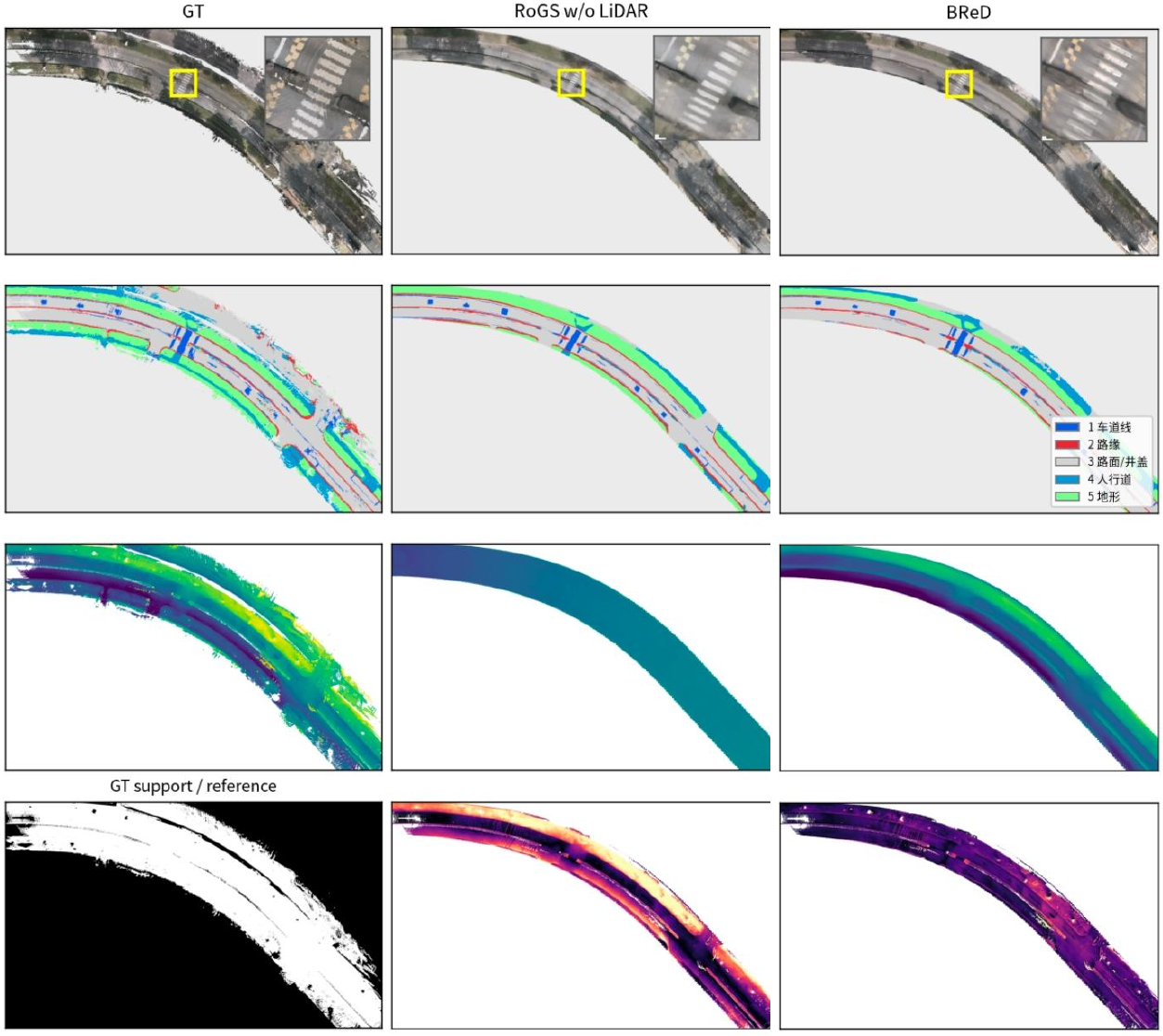}
  \caption{\textbf{Reconstruction at a complex curved intersection.}
  Columns show GT, RoGS w/o LiDAR, and BReD; rows show BEV RGB, semantics, elevation, and elevation residual over the GT support region. Darker residuals indicate higher consistency with GT.}
  \label{fig:extreme_scenarios}
\end{figure}

\vspace{-3pt}
\noindent\textbf{Reconstruction Quality over Extreme Scenarios.}
Fig.~\ref{fig:extreme_scenarios} compares reconstruction at a curved intersection with dense crosswalks and multiple camera seams. Both methods recover comparable RGB appearance and semantic structures, including crosswalks, lane lines, road edges, and sidewalks. However, RoGS w/o LiDAR produces an almost uniform surface with limited transverse variation, whereas BReD preserves distinct elevations for the lane crown, curb, and sidewalk. Accordingly, BReD produces sparse residuals mainly near object boundaries, while RoGS shows widespread errors along the outer road. This comparison attributes BReD's advantage in this challenging scene primarily to elevation accuracy rather than photometric or semantic quality.
\begin{figure}[htp]
  \centering
  \includegraphics[width=0.9\linewidth]{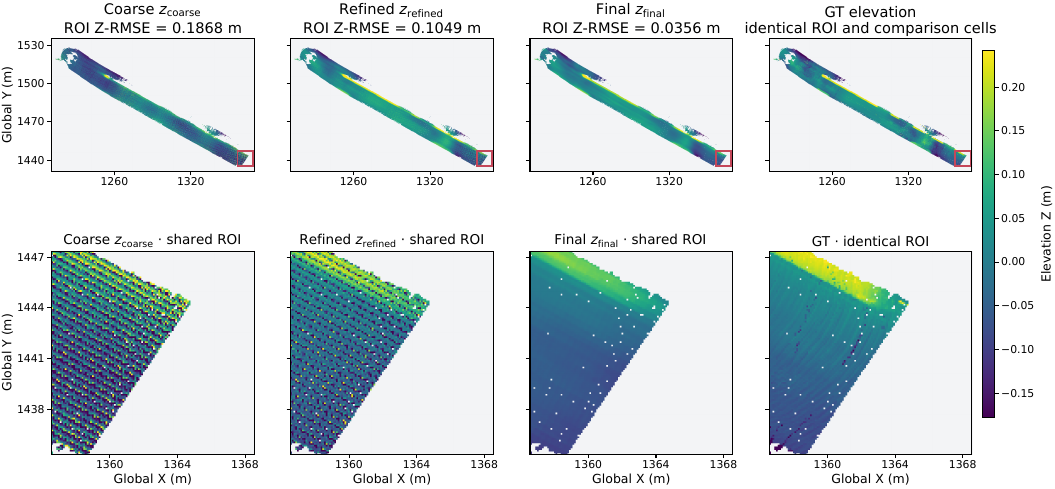} 
  \caption{\textbf{Progressive elevation evolution.} 
  Top: Holistic elevation maps. Bottom: Zoomed-in $12\text{m} \times 12\text{m}$ ROI. Visually, the surface transitions from $z_{\text{coarse}}$ to $z_{\text{refined}}$, and ultimately to $z_{\text{final}}$.}
  \label{fig:elev_evolution}
\end{figure}

\noindent\textbf{Progressive Elevation Evolution.} 
To illustrate how our pipeline progressively refines road geometry, Fig.~\ref{fig:elev_evolution} visualizes the stage-wise elevation evolution. The initial camera-derived $z_{\text{coarse}}$ exhibits severe discretization artifacts (local ROI RMSE: $18.68\text{cm}$). \texttt{ReD*} then suppresses this spatial noise to recover a continuous macroscopic slope in $z_{\text{refined}}$ ($10.49\text{cm}$). Finally, initialized with $z_{\text{refined}}$, \texttt{FFGS} sharpens fine-grained boundaries in $z_{\text{final}}$ ($3.56\text{cm}$). Beyond numerical gains, this progression explicitly demonstrates the resolution of jagged depth ambiguities from initial view transformations, yielding a physically plausible topography.

\begin{figure}[t]
  \centering
  \includegraphics[width=0.9\linewidth]{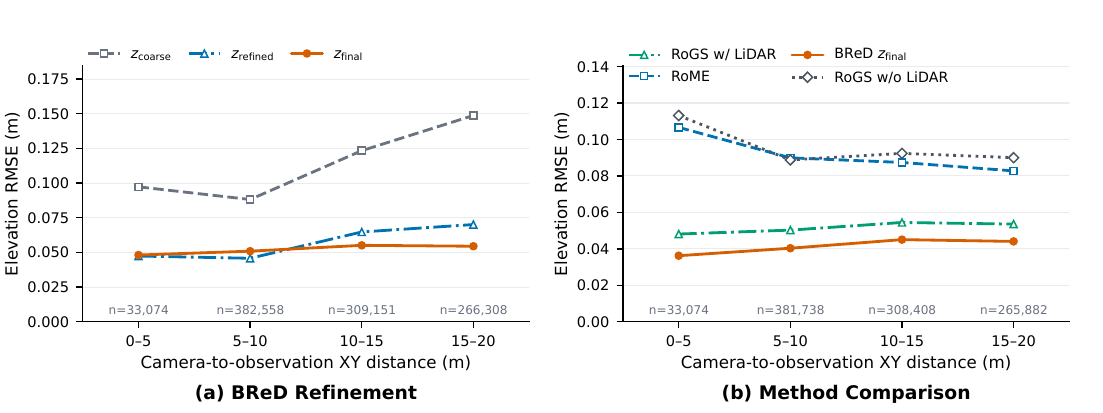} 
  \caption{\textbf{Elevation RMSE versus camera-to-observation XY distance.} 
  (a) Stage-wise progression showing error reduction from $z_{\text{coarse}}$ to refined stages. 
  (b) Comparison against state-of-the-art baselines.}
  \label{fig:distance_error}
\end{figure}
\noindent\textbf{Elevation Error vs. Camera Distance.}
We evaluate vertical RMSE over camera-to-observation XY distances of $0\text{--}20\,\mathrm{m}$, using $5\,\mathrm{m}$ bins with valid counts $n$ reported in Fig.~\ref{fig:distance_error}. As shown in Fig.~\ref{fig:distance_error}(a), our hierarchical pipeline consistently reduces the large errors of $z_{\text{coarse}}$. Although $z_{\text{final}}$ is slightly worse than $z_{\text{refined}}$ at $5\text{--}10\,\mathrm{m}$, it achieves substantially lower errors at $10\text{--}20\,\mathrm{m}$, demonstrating its effectiveness for challenging distant predictions. This advantage is also qualitatively evident in Fig.~\ref{fig:elevation_projection}, where $z_{\text{final}}$ exhibits less extreme error colors for distant projected points. Moreover, Fig.~\ref{fig:distance_error}(b) shows that our fully feed-forward, camera-only inference BReD ($z_{\text{final}}$) achieves the lowest error across all distance bins.
\vspace{-3pt}
\section{Conclusion}
\vspace{-3pt}
We presented \textbf{BReD}, a feed-forward framework for sequence-level road reconstruction from calibrated surround-view cameras. BReD combines LiDAR-supervised BEV lifting during training, single-step residual denoising, temporal surface aggregation, and geometry-prior-guided Gaussian decoding. It uses camera-only inference without scene-specific optimization and achieves $7.34\,\mathrm{cm}$ elevation RMSE and $90.92\%$ mIoU on four held-out nuScenes scenes, providing an effective alternative to per-scene road reconstruction.

\vspace{-3pt}
\section*{Use of Large Language Models}
\vspace{-3pt}
GPT-5.6 Sol and Gemini Argon were used only for grammar and wording improvement; all content is the authors' own.
{
    \small
    \bibliographystyle{ieeenat_fullname}
    \bibliography{main}
}


\end{document}